# Neural Networks for Handwritten English Alphabet Recognition

Yusuf Perwej
Department of Computer Science
Singhania University
Rajsthan, India

Ashish Chaturvedi
Department of Computer Science
Kishan Institute of Engg. & Technology
Meerut, India

## ABSTRACT
This paper demonstrates the use of neural networks for developing a system that can recognize hand-written English alphabets. In this system, each English alphabet is represented by binary values that are used as input to a simple feature extraction system, whose output is fed to our neural network system.

## 1. INTRODUCTION
Optical character recognition is the past when in 1929 Gustav Tauschek got a patent on OCR in Germany followed by Handel who obtained a US Patent on OCR in USA in 1933. Since then number of character recognition systems have been developed and are in use for even commercial purposes also. But still there is a hope to build some more intelligent hand written character recognition system because hand writing differ from one person to other. His writing style, shape of alphabets and their sizes makes the difference and complexity to recognize the characters.

Researchers already paid many efforts in designing hand written character recognition system most of them cited as [1-5] because of its important application like bank checking process, reading postal codes and reading different forms [6]. Handwritten digit recognition is still a problem for many languages like Arabic, Farsi, Chinese, English, etc [7]. A machine can perform more tasks than a human being in the same time; this kind of application saves time and money and eliminates the requirement that a human perform such a repetitive task. For the recognition of English handwritten characters, various methods have been proposed [8-12]. Also a few numbers of studies have been reported for Farsi language [13-15].

In some hand-writing, the characters are indistinguishable even to the human eye, and that they can only be distinguished by context. In order to distinguish between such similar characters, the tiny differences that they have must be identified. One of the major problems of doing this for hand written characters is that they do not appear at the same relative location of the letter due to the different proportions in which characters are written by different writers of the language. Even the same person may not always write the same letter with the same proportions.

Here, the goal of a character recognition system is to transform a hand written text document on paper into a digital format that can be manipulated by word processor software. The system is required to identify a given input character form by mapping it to a single character in a given character set. Each hand written character is split into a number of segments (depending on the complexity of the alphabet involved) and each segment is handled by a set of purpose built neural network. The final output is unified via a lookup table. Neural network architecture is designed for different values of the network parameters like the number of layers, number of neurons in each layer, the initial values of weights, the training coefficient and the tolerance of the correctness. The optimal selection of these network parameters certainly depends on the complexity of the alphabet.

## 2. HAND WRITTEN ENGLISH ALPHABET RECOGNITION SYSTEM
Two phase processes are involved in the overall processing of our proposed scheme: the Pre-processing and Neural network based Recognizing tasks. The pre-processing steps handle the manipulations necessary for the preparation of the characters for feeding as input to the neural network system. First, the required character or part of characters needs to be extracted from the pictorial representation. The splitting of alphabets into 25 segment grids, scaling the segments so split to a standard size and thinning the resultant character segments to obtain skeletal patterns. The following pre-processing steps may also be required to furnish the recognition process:

I. The alphabets can be thinned and their skeletons obtained using well-known image processing techniques, before extracting their binary forms.

II. The scanned documents can be "cleaned" and "smoothed" with the help of image processing techniques for better performance.

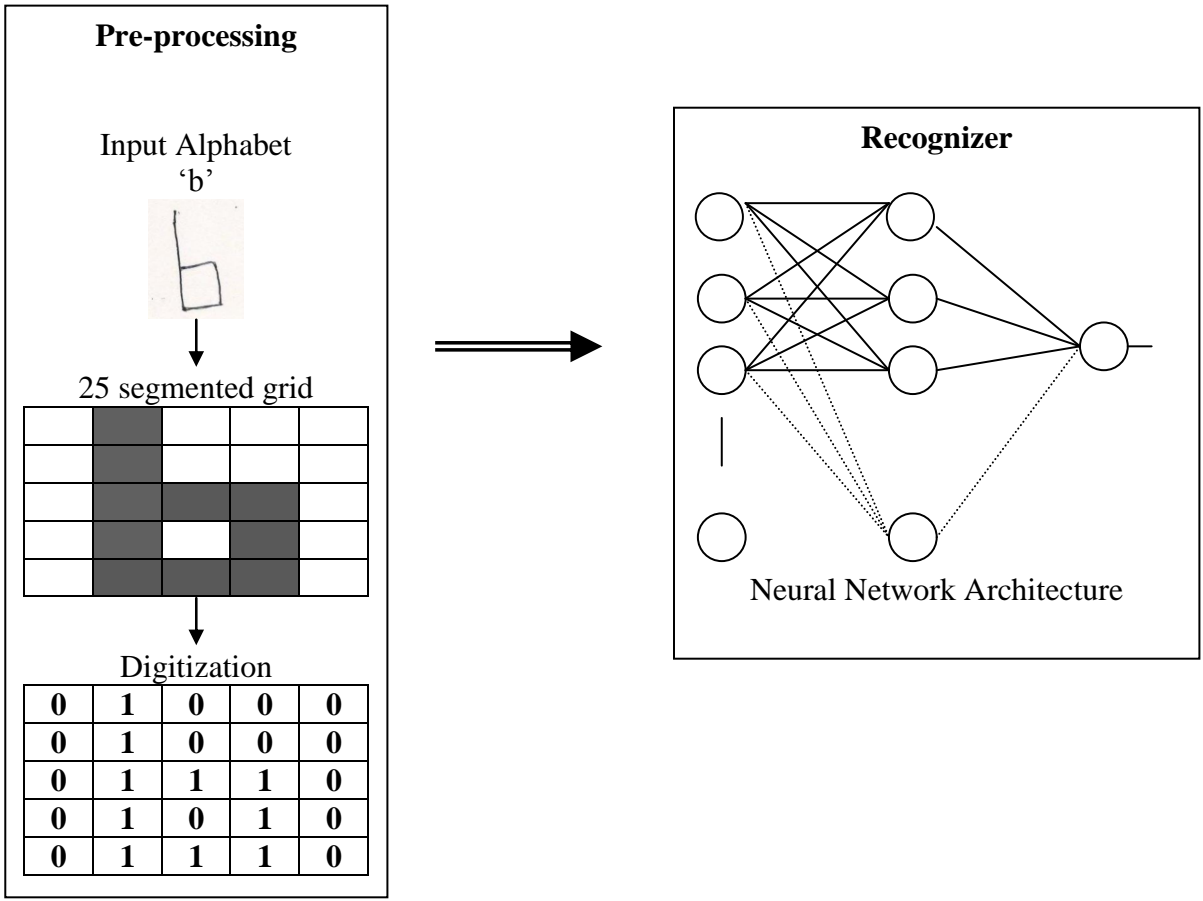

**Figure 1: Two Phase Character Recognizer**

Further, the step involves the digitization of the segment grids because neural networks need their inputs to be in the form of binary (0's & 1's). The next phase is to recognize segments of the character. This is carried out by neural networks having different network parameters. The each digitize segment out of 25 segmented grid is then provided as input to the each node of neural network designed specially for the training of that segments. Once the networks trained for these segments, be able to recognize them. Characters which are similar looking but distinct are also distinguished at this time. Results obtained were satisfactory, especially when input characters were very close to printed letters.

## 3. RESULTS & DISCUSSIONS

The proposed alphabet recognition system was trained to recognize hand written English alphabets. Since the alphabets are divided into 25 segments, neural network architecture is designed specially for the processing of 25 input bits. The network parameters used for training are:

Learning rate coefficient = 0.05

No. of Units in Input layer = 25

No. of Hidden Layers = 2

No. of Units in Hidden layer = 25

Initial Weights = Random [0,1]

Transfer Function Used for Hidden Layer 1 = "Logsig"

Transfer Function Used for Hidden Layer 2 = "Tansig"

The training set involves the binary codes of alphabets. It was not practical to input these shapes individually when creating training sets, because the shape of a particular segment of the actual character depends on handwriting. Therefore, this was automated so that the entire letter is input to the system, and then the shape of the segment needed is extracted from this full letter instead of drawing the shape of the segment itself. The examples of training set patterns can be seen below:

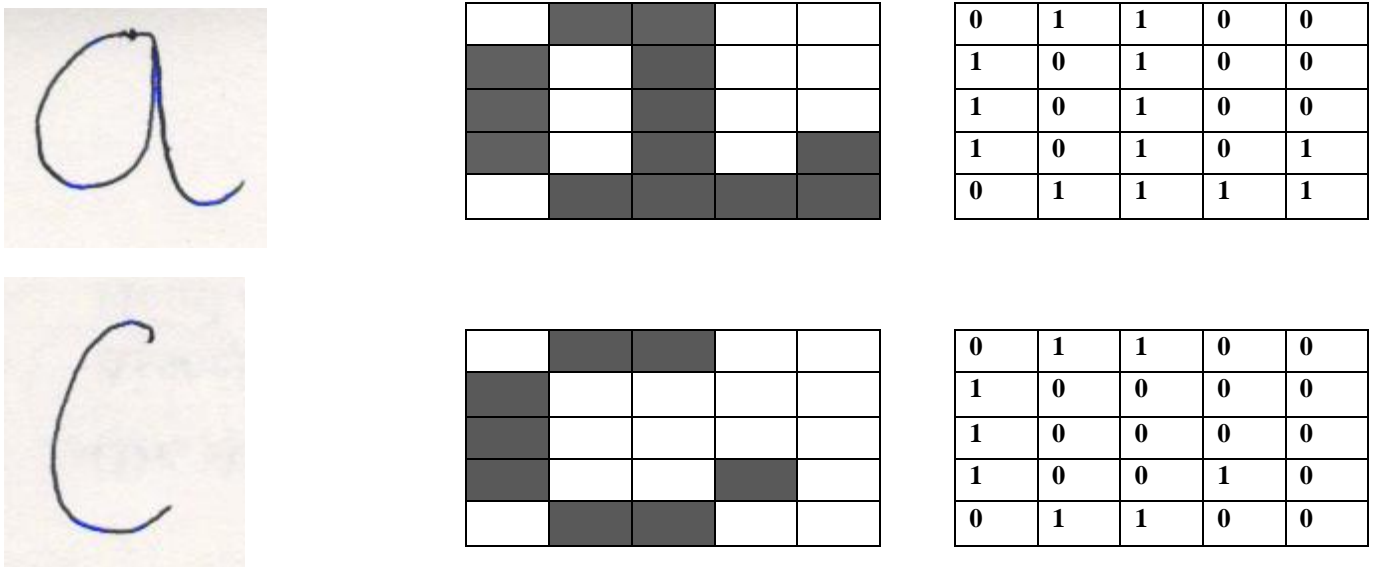

**Figure 2: Example Training sets**

The binary input of each alphabet of different hand writing styles is then feed to all the units of input layer of the network at once and the network after processing through hidden layers and with the training algorithm, Gradient Descent Back propagation, can be trained for these inputs. Now, the trained network is capable to recognize the alphabet of different hand writing. The examples of test set patterns can be as below:

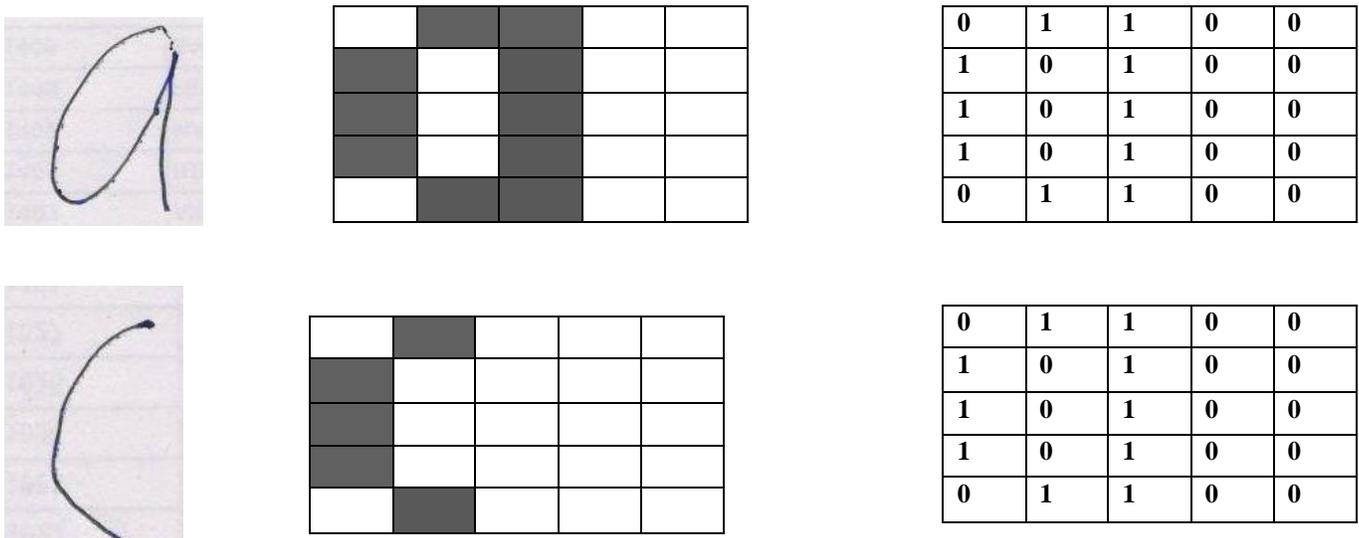

**Figure 3: Example Testing Set**

The result can be tabulated as:

| Alphabet | No. of samples for training | No. of samples for testing | No. of epochs | % Recognition Accuracy |
|---|---|---|---|---|
| a | 20 | 5 | 294 | 94.0 |
| b | 20 | 5 | 321 | 83.0 |
| c | 20 | 5 | 587 | 71.0 |
| d | 20 | 5 | 282 | 88.0 |
| e | 20 | 5 | 548 | 64.0 |
| f | 20 | 5 | 254 | 85.0 |
| g | 20 | 5 | 247 | 89.0 |
| h | 20 | 5 | 263 | 92.0 |
| i | 20 | 5 | 658 | 72.0 |
| j | 20 | 5 | 599 | 73.0 |
| k | 20 | 5 | 300 | 91.0 |
| l | 20 | 5 | 652 | 71.0 |
| m | 20 | 5 | 456 | 86.0 |
| n | 20 | 5 | 398 | 82.0 |
| o | 20 | 5 | 356 | 94.0 |
| p | 20 | 5 | 264 | 88.0 |
| q | 20 | 5 | 287 | 82.0 |
| r | 20 | 5 | 669 | 70.0 |
| s | 20 | 5 | 202 | 88.0 |
| t | 20 | 5 | 252 | 79.0 |
| u | 20 | 5 | 458 | 80.0 |
| v | 20 | 5 | 488 | 77.0 |
| w | 20 | 5 | 511 | 94.0 |
| x | 20 | 5 | 341 | 91.0 |
| y | 20 | 5 | 268 | 71.0 |
| z | 20 | 5 | 296 | 90.0 |

We can observe from the table that recognition accuracy is lower for the similar input patterns like: c & e; i, j, l & r; and u & v. for these similar patterns in different hand writing, even human eye can not be able easily distinguish, hence the machine needs lots of training epochs to recognize them but with some misclassifications.

## 4. CONCLUSION

We have proposed and developed a scheme for recognizing hand written English alphabets. We have tested our experiment over all English alphabets with several Hand writing styles. Experimental results shown that the machine has successfully recognized the alphabets with the average accuracy of 82.5%, which significant and may be acceptable in some applications. The machine found less accurate to classify similar alphabets and in future this misclassification of the similar patterns may improve and further a similar experiment can be tested over a large data set and with some other optimized networks parameters to improve the accuracy of the machine.